\documentclass{article}
 \usepackage[nonatbib, preprint]{neurips_2026}
\usepackage{float}
\usepackage{xcolor}
\usepackage{colortbl}
\newcommand{\tok}[1]{{\scriptsize\color{gray}(#1)}}
\usepackage{amsmath,amsfonts,amssymb}
\usepackage{algpseudocode}
\usepackage{algorithm}
\usepackage{array}
\usepackage[caption=false,font=normalsize,labelfont=sf,textfont=sf]{subfig}
\usepackage{textcomp}
\usepackage{stfloats}
\usepackage{url}
\usepackage{verbatim}
\usepackage{booktabs}
\usepackage{graphicx}
\usepackage[colorlinks=true,
            linkcolor=red,
            citecolor=blue
            ]{hyperref}
\usepackage{multirow}
\usepackage[T1]{fontenc}
\usepackage{nicefrac}       
\usepackage{microtype}      
\usepackage{xcolor}         
\usepackage{float}
\usepackage{lscape}
\usepackage{wrapfig}
\usepackage{subcaption}
\usepackage{multicol}
\usepackage{tikz}
\usepackage{bm}
\usepackage{svg}
\usetikzlibrary{arrows.meta, positioning}
\usepackage{tcolorbox}
\usepackage{natbib}

\definecolor{steer}{RGB}{210,230,255}
\definecolor{steerblue}{RGB}{31,58,104}
\definecolor{basegold}{RGB}{154,116,20}
\definecolor{cotgray}{RGB}{91,91,88}
\newcommand{\rowtag}[2]{\colorbox{#1}{\textcolor{white}{\sffamily\bfseries\tiny #2}}}
\newcommand{\ntok}[1]{{\scriptsize\color{gray}#1 tokens}}
\newcommand{\okmark}{\textcolor{steerblue}{\checkmark}}
\newcommand{\badmark}{\textcolor{basegold}{$\times$}}

\title{COMPASS: Finding Where Reasoning Lives in Language Models}
\author{%
  Pratyay Dutta\thanks{University of California, Riverside, CA 92521.
    Work performed during a summer internship at Lawrence Livermore National Laboratory} \\
  \texttt{pdutt005@ucr.edu} \\
  \And
  Kowshik Thopalli\thanks{Lawrence Livermore National
    Laboratory, Livermore, CA 94550.}\\
  \texttt{thopalli1@llnl.gov} \\
  \And
  Vivek Narayanaswamy\footnotemark[2]  \\
  \texttt{narayanaswam1@llnl.gov} \\
}
\date{August 2026}

\begin{document}
\maketitle

\begin{abstract}
Explicitly eliciting reasoning substantially improves LLM performance. Existing approaches require a predefined characterization of reasoning, whether through CoT prompt design, contrastive CoT directions, or via SAE derived reasoning features. For mathematical reasoning with verifiable answers, we show that a much simpler signal suffices, which is the correctness of the model's own direct answer attempts. This signal yields a latent direction that elicits reasoning. This direction is decodable within the activations of most attention heads, but only a small subset of them can be effectively intervened. We introduce COMPASS, an inference-time steering method that identifies these heads using a logit-space attribution score and steers their activations along the correctness direction, requiring only per-head activation statistics. Across three model families and multiple math benchmarks, COMPASS outperforms the activation-steering baselines we compare against, improves GSM8K accuracy by 16 percentage points on average, and approaches CoT accuracy with 20-70\% fewer generated tokens. Interventions transfer without re-fitting to unseen benchmarks, and ablations show that both the correctness direction and the small set of heads carrying it are necessary, with the effect concentrated in remarkably few heads. 

\end{abstract}

\section{Introduction}
\label{sec:intro}

Large language models (LLMs) exhibit strong reasoning capabilities across a wide range of tasks~\cite{kojima2022large, hendrycks2021math}. A standard approach for eliciting reasoning is chain-of-thought (CoT) prompting~\cite{wei2022chain}, which encourages the model to generate intermediate steps before producing a final answer. Recent work, however, suggests that reasoning-relevant computation also resides in the model's hidden states and can be elicited without explicit CoT prompting~\cite{he2026reasoning, sun2026llm, deng2023implicit}, indicating that internal representations contain meaningful computational signatures that can be intervened upon.

For instance, Sparse Autoencoders (SAEs) have been used to elicit reasoning by identifying a reasoning-associated feature, though doing so requires a contrast condition prompt and a separate SAE per layer~\cite{he2026reasoning}. More broadly, linear probes, including probes trained on individual attention head activations, have been shown to decode or steer behavioral properties such as truthfulness from internal states. However only such specific properties have been steered so far, not reasoning~\cite{li2023iti, vonrutte24a, ashok2026language, li2026manifoldguided}. Each of these approaches requires some predefined characterization of reasoning, whether through a contrast prompt, a trained auxiliary model, or a property specific probe. For mathematical reasoning with verifiable answers, which is our focus, whether an intervention can be found without additional requirements and specifications remains underexplored.

Surprisingly, we find that an effective reasoning intervention can be elicited from a signal as simple as a direction separating correctly and incorrectly answered samples, requiring no CoT demonstrations, reasoning-specific information, or contrastive direct-versus-CoT prompting. This direction is estimated locally, at each attention head, by using last token activations of the brief input prompts alone associated with the correct and incorrect generations.

The key remaining challenge is where to intervene. Correctness information is decodable across many attention heads. However, this decodability alone is not causally sufficient for effective steering~\cite{vonrutte24a, li2023iti}. Heads selected without regard to causal effect, including at random, yield no reliable benefit (Fig.~\ref{fig:ablation-heads}, left). We find that only a small subset of heads are thus useful for reasoning intervention. To that end, we introduce COMPASS (\textbf{CO}rrectness-\textbf{M}apped \textbf{P}er-head \textbf{A}ttribution for \textbf{S}teering), an inference-time steering framework that identifies and steers the heads with the highest output sensitivity to the correctness direction, rather than selecting heads by decodability alone. Unlike SAE-based approaches, COMPASS needs no trained auxiliary model, only the per-head mean activation computed separately over correct and incorrect samples, making the method as cheap to deploy as it is effective.

Remarkably, COMPASS is never provided with a representation of reasoning itself, yet steering the selected heads elicits concise intermediate computations that often convert incorrect direct answers into correct solutions. Across three model families~\cite{grattafiori2024llama3,yang2025qwen3,gemmateam2026gemma4} and multiple mathematical reasoning benchmarks~\cite{cobbe2021gsm8k,patel2021svamp,hendrycks2021math}, COMPASS outperforms existing activation-steering baselines in most model--benchmark pairs, improves GSM8K accuracy by 16 points on average, and approaches CoT performance with substantially shorter generations. Ablations show that both the correctness direction and its intervention location are necessary, with much of the \emph{effect concentrated in only a few heads}. The identified interventions further transfer without re-fitting to GSM-Plus~\cite{li2024gsmplus} and HARP~\cite{yue2024harp}.

Our main contributions are:
\begin{itemize}
    \item \textbf{Reasoning intervention from outcome supervision alone.}
    We show that correct/incorrect outcomes help recover an effective reasoning intervention, without CoT demonstrations, predefined reasoning directions, or identified reasoning features.

    \item \textbf{Attribution-based steering.}
    We introduce COMPASS, which combines per-head correctness directions with their predicted downstream logit effect to identify effective intervention sites.

    \item \textbf{Concentrated internal control.}
    Controlled head and direction ablations show that both, what is injected and where it is injected matter, with much of the intervention effect concentrated in remarkably few attention heads.

    \item \textbf{Accuracy, efficiency, and transfer.}
    Across three model families and multiple reasoning benchmarks, COMPASS closes much of the gap to explicit CoT, with substantially shorter generations and without any re-fitting on unseen distributions, demonstrating that the same lightweight intervention is simultaneously accurate, efficient, and general.
\end{itemize}

\section{Related Work}
\label{sec:background}

\noindent \textbf{Eliciting Reasoning from LLMs.}
The reasoning capabilities of LLMs are most commonly elicited through chain-of-thought (CoT) prompting~\cite{wei2022chain, kojima2022large}, where performance improves significantly as the model is made to think through intermediate steps before answering. However, prompting is not the only route to reasoning. Recent work shows that the internal states of the LLM carry useful computational signatures that can be extracted directly, without relying on the prompt alone. Fractional reasoning~\cite{liu2025fractional} takes a direct prompt and a CoT prompt, identifies the linear direction separating their latent activations, and shifts the model into a controlled reasoning regime along this direction. In a similar spirit, sparse autoencoder features extracted from direct and CoT prompts have been used to uncover a latent reasoning mode~\cite{he2026reasoning}, and contrastive prompt pairs have been used to steer metacognitive behaviors in ~\cite{li2026metacognition}. Each of these methods requires some prior specification of what reasoning should look like, rather than discovering it directly from the model's own behavior. In contrast, COMPASS requires no such explicit specifications or requirements.

\noindent \textbf{Probing Hidden States of LLMs.}
A second line of work extracts information from hidden states via probing, using simple classifiers, typically linear, trained on activations to predict a target property. Inference-time intervention \cite{li2023iti} trains such probes at every attention head and shifts activations along the probe direction at the heads where properties (e.g., truthfulness) are most detectable. Follow-up work refined how the probe guides generation~\cite{vonrutte24a, li2026manifoldguided}, and applied probing to new targets, including whether a model believes its own answer is correct~\cite{zhang2025selfverification}, whether a math problem is solvable at all~\cite{xiros2026solvability}, and whether code is correct before it is written~\cite{dicicco2026code}. However, each method trains a probe on a single, curated property, such as truthfulness, honesty, or solvability, requiring its own labeled dataset, and only a subset of these properties has been shown to be steerable at all and none for reasoning. Our steering or intervention direction requires only a labeled set of single-pass generations, split by the LLM's own correctness, with heads selected by their causal effect on the output rather than by property detectability alone.

\noindent \textbf{Activation Steering in LLMs.}
A third line of work steers behavior directly through activations rather than through the prompt. This traces back to in-context learning~\cite{li2024language, dherin2025learning}, where prompt demonstrations change model behavior by shifting internal activations, an effect shown to be compressible into a single latent task vector~\cite{hendel2023task}. In-context vectors made this actionable by distilling demonstrations into a vector added back during inference~\cite{liu2024incontext}. Subsequent methods derived directions from paired prompts~\cite{turner2023actadd, rimsky2024caa} or curated datasets~\cite{zou2023repe}, applied steering conditionally on the input~\cite{lee2025cast}, synthesized query specific vectors from a mixture of steering experts~\cite{weng2026finesteer}, or restricted intervention to an invariant low rank subspace of reasoning trajectories~\cite{malarkkan2026tilr}. Each requires paired task demonstrations, a separate forward pass, or a labeling scheme. Our steering direction requires none of these, extracted instead from the model's own outcome labels in a single pass.

\section{Proposed Approach}
\label{sec:approach}

In this section, we present COMPASS, a method that recovers a coarse correctness direction from a model's own hidden states, identifies the attention heads that causally carry this direction to the output, and steers only those heads to elicit reasoning.

Our key insight builds on the observations of~\cite{li2023iti, vonrutte24a} that behavioral properties such as truthfulness or honesty, properties that at a high level resemble correctness, are decodable from the activations of several attention heads. However, only a small subset of these heads causally affect the output, which we define as a head whose intervened activation changes the model's generated answer, as opposed to merely correlating with it. We introduce a selection procedure that distinguishes these causal heads from the many that merely encode correctness as a decodable signal. We find that the causal effect is concentrated in remarkably few heads. 

COMPASS answers two questions keeping the LLM frozen: (i) which direction in activation space separates the model's correct attempts from its incorrect ones (\S\ref{sec:direction}), and (ii) which attention heads carry this direction into the output when it is intervened upon (\S\ref{sec:heads}).

\noindent \textbf{Preliminaries.} Consider a decoder-only transformer with $L$ layers and vocabulary $\mathcal{V}$. Each layer has $H$ attention heads, hidden dimension $d$, and per-head dimension $d_h = d/H$. Let $x_l \in \mathbb{R}^d$ denote the layer $l$ input at a fixed token position. Head $(l,h)$ computes an attention weighted value vector $a_l^h \in \mathbb{R}^{d_h}$, which is projected into the layer output through its slice $W_O^{l,h} \in \mathbb{R}^{d \times d_h}$ of the corresponding output projection matrix. The next layer input is updated as
\begin{equation}
x_{l+1} \;=\; x_l \;+\; \sum_{h=1}^{H} W_O^{l,h}\,a_l^h \;+\; \mathrm{mlp}_l,
\label{eq:resid}
\end{equation}
where $\mathrm{mlp}_l$ is the output of the layer's MLP sublayer. After the final layer $L$, $x_{L+1}$ is normalized and mapped to a next token distribution by the unembedding matrix $W_U \in \mathbb{R}^{|\mathcal{V}| \times d}$,
\begin{equation}
p \;=\; \mathrm{softmax}(W_U\,\mathrm{LN}(x_{L+1})) \;\in\; \mathbb{R}^{|\mathcal{V}|},
\label{eq:unembed}
\end{equation}
where $\mathrm{LN}$ denotes the LayerNorm.

\subsection{Estimating the Correctness Direction}
\label{sec:direction}
Let $\mathcal{D} = \{(q_i, y_i^\star)\}_{i=1}^N$ be a training set of problems $q_i$ with ground truth answers $y_i^\star$. For every $q_i$, we run a fixed direct answer prompt once and obtain the generation $g_i$. We cache the activation $a_l^h(q_i)$ of every attention head $(l,h)$ at the last prompt token position, before generation begins. Each generation $g_i$ is graded against $y_i^\star$, yielding a binary label $y_i \in \{0,1\}$, where $y_i = 1$ if $g_i$ is correct and $y_i = 0$ otherwise. The label $y_i$ indicates whether the model, under the direct answer prompt, answers $q_i$ correctly.


Following~\cite{li2023iti}, for head $(l,h)$ we estimate a correctness direction $\theta_l^h$ as the unit vector along the difference between the mean activation on correct samples and the mean activation on incorrect samples. This direction captures the axis along which the two populations tend to shift, on average, and is defined as
\begin{equation}
\mu_{+}^{l,h}=\mathbb{E}\,[a_l^h \mid y{=}1],\quad
\mu_{-}^{l,h}=\mathbb{E}\,[a_l^h \mid y{=}0],\quad
\theta_l^h=\frac{\mu_{+}^{l,h}-\mu_{-}^{l,h}}{\|\mu_{+}^{l,h}-\mu_{-}^{l,h}\|}.
\label{eq:theta}
\end{equation}
As a unit vector, $\theta_l^h$ specifies the direction of the correctness axis but not the scale of the head's activations along it. A fixed shift magnitude is not comparable across heads, since different heads have different activation scales. We therefore normalize the shift using the standard deviation of each head's activations along its own correctness direction,
\begin{equation}
\sigma_l^h = \operatorname{std}_{q \in \mathcal{D}}\big(\langle a_l^h(q), \theta_l^h \rangle\big).
\label{eq:sigma}
\end{equation}
A displacement of $\alpha\,\sigma_l^h$ along $\theta_l^h$ corresponds to $\alpha$ standard deviations along the correctness direction, making $\alpha$ a scale invariant steering strength shared across all heads.

\begin{figure}[t]
\centering
\includegraphics[width=1.0\linewidth]{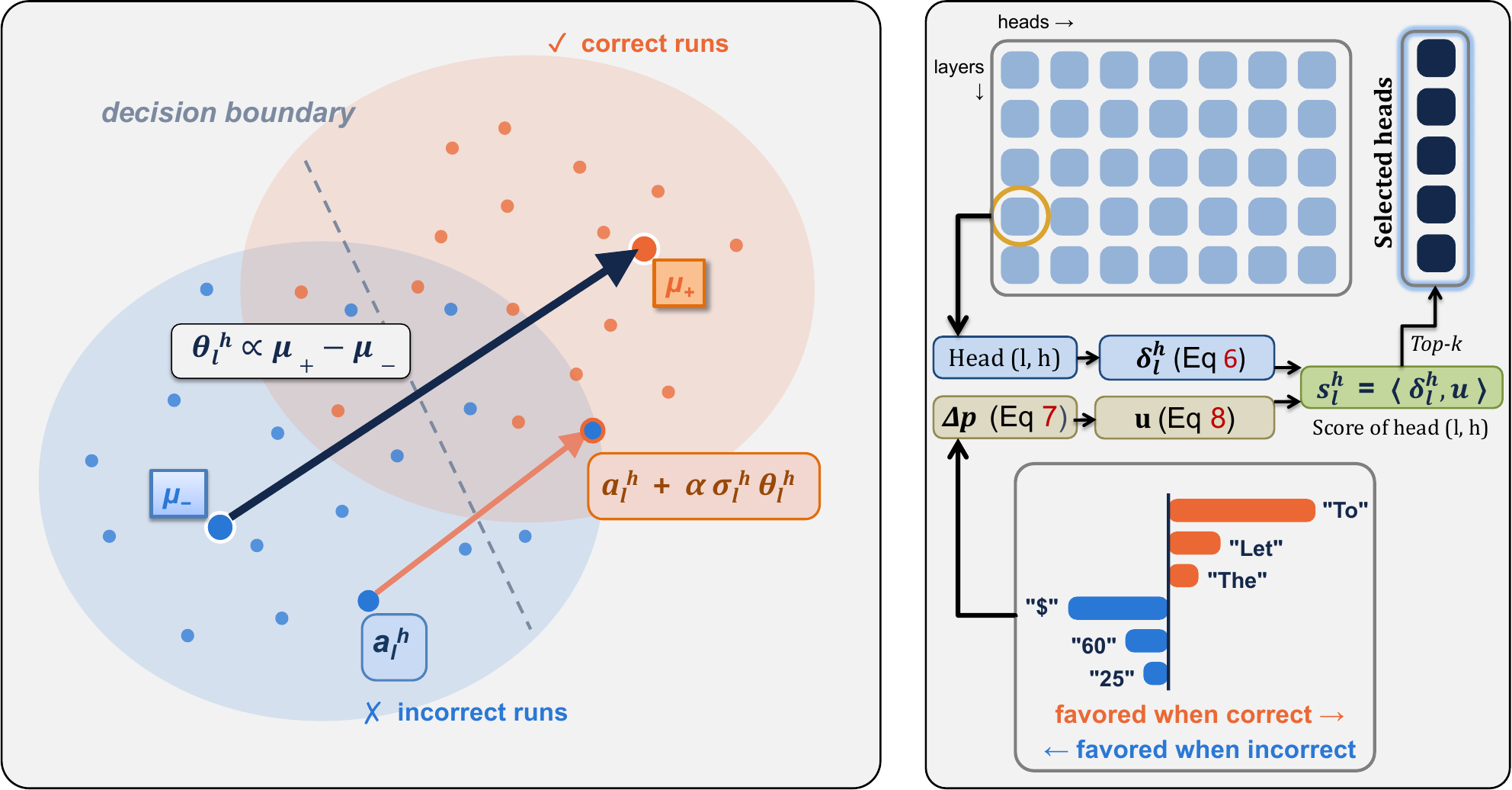}
\caption{\textbf{Left: The per-head correctness direction.} In the activation space of a head, activations from correctly and incorrectly answered problems form two populations, with means $\mu_+^{l,h}$ and $\mu_-^{l,h}$. $\theta_l^h$ is the unit vector connecting these means, and $\sigma_l^h$ is the standard deviation of the activations projected onto $\theta_l^h$. Steering shifts an activation on the incorrect side by $\alpha\,\sigma_l^h$ along $\theta_l^h$. \textbf{Right: Head selection by attribution score.} Top: the score $s_l^h$ of each head, with the $K$ heads selected within the promoted layer band outlined in blue. Bottom: the outcome contrast $\Delta p$ measured on the fitting rows. First generated tokens for correct answers concentrate on ``To'', and for incorrect answers on ``\$''.}
\label{fig:com-heads}
\end{figure}

Following the intervention procedure of ITI~\cite{li2023iti}, we steer the model by adding $\alpha\,\sigma_l^h\,\theta_l^h$ to the output of each selected head, before it is projected into the layer input,
\begin{equation}
x_{l+1} \;=\; x_l \;+\; \sum_{h=1}^{H} W_O^{l,h}\big(a_l^h+\alpha\,\sigma_l^h\,\theta_l^h\big) \;+\; \mathrm{mlp}_l,
\label{eq:iti}
\end{equation}
where $\theta_l^h \equiv 0$ for every head not selected for steering. Since $W_O^{l,h}$ is linear, each unit of $\alpha$ contributes a fixed vector to the layer input,
\begin{equation}
\delta_l^h \;=\; W_O^{l,h}\,\sigma_l^h\,\theta_l^h\;\in\;\mathbb{R}^{d},
\label{eq:delta}
\end{equation}
which is the correctness direction of head $(l,h)$, projected into the layer input. $\delta_l^h$ is the contribution head $(l,h)$ makes to the layer input per unit of $\alpha$. 

\subsection{Where to Intervene: Selecting Causal Heads}
\label{sec:heads}
Equation~\eqref{eq:iti} leaves the choice of heads open. The direction $\theta_l^h$ from Eq.~\eqref{eq:theta} exists for every head, since a difference of means can be computed regardless of whether that head's output has any effect on the model's generation. Decodability, in this sense, only requires that correct and incorrect activations differ on average, it places no constraint on whether adding $\delta_l^h$ to the layer input changes the final output. Selecting heads by decodability alone can therefore result in heads with no causal effect, or with an effect that opposes the intended direction. This is confirmed empirically in the ITI row of Table~\ref{tab:main}, where steering the heads that decode correctness best yields little or no improvement, and is at times detrimental. On MATH-500, it lowers Llama-3.1-8B accuracy from 40.8\% to 23.6\%, and Qwen3-4B from 53.8\% to 30.6\%.

A head is useful for intervention only if its contribution $\delta_l^h$ changes the model's output. Decodability in activation space, as used to define $\theta_l^h$, only tells us that correct and incorrect activations differ, not whether that difference affects the generated token. To capture causal effect, the selection criterion must therefore be defined in terms of the shift a head's contribution produces in the next token logits, rather than in the space of activations. The unembedding matrix $W_U$ maps a vector in the layer input space to a vector of next token logits (Eq.~\ref{eq:unembed}). Rather than mapping $\delta_l^h$ forward through $W_U$, which would require rerunning the model, we use the identity $\langle W_U x, z \rangle = \langle x, W_U^\top z \rangle$. For any target logit direction $z$, its inner product with a layer input vector $x$ can be computed entirely in layer input space, using $W_U^\top z$. This lets us score $\delta_l^h$ directly against a target logit direction, without any additional forward pass. 

We first define a target shift in the space of first generated tokens. Following~\cite{zhao2024firsttoken}, who show that an LLM's knowledge of whether its answer will be correct is most concentrated in the distribution of its first generated token, we use this distribution as our target. Let $p_+, p_- \in \Delta^{|\mathcal{V}|}$ be the empirical first generated token distributions for questions the model answers correctly and incorrectly, under the same prompt. Their difference,
\begin{equation}
\Delta p \;=\; p_+ - p_-,
\label{eq:dp}
\end{equation}
is positive on tokens more likely to precede a correct answer and negative on tokens more likely to precede an incorrect answer (Figure~\ref{fig:com-heads}(b)).

$\Delta p$ lies in vocabulary space, while $\delta_l^h$ lies in layer input space. To compare them, we map $\Delta p$ into layer input space using the transpose of the unembedding matrix,
\begin{equation}
u \;=\; W_U^{\top}\,\Delta p \;\in\; \mathbb{R}^{d}.
\label{eq:u}
\end{equation}
We call $u$ the target direction. A head whose contribution aligns with $u$ shifts the first generated token logits toward tokens that precede correct answers.

We score head $(l,h)$ by how well its contribution $\delta_l^h$ aligns with the target direction $u$,
\begin{equation}
s_l^h \;=\; \langle u, \delta_l^h \rangle \;=\; W_U^{\top}\,\Delta p\, W_O^{l,h}\,\sigma_l^h\,\theta_l^h.
\label{eq:score}
\end{equation}
A high score implies that $\delta_l^h$ is large and points in the direction of $u$. We select the top $K$ heads by $s_l^h$ and steer only these heads, setting $\theta_l^h \equiv 0$ for all other heads.

Across three open weight models and multiple short answer math benchmarks, intervening only these selected heads consistently improves accuracy over standard prompt generation. This requires a single forward pass over the training dataset, and steering vectors are applied zero shot at inference. Algorithm~\ref{alg:compass} summarizes the steering procedure.

Figure~\ref{fig:examples} shows this effect on two problems. In both cases, the direct answer prompt produces an incorrect answer. The steered model instead opens with "To", derives the answer in a few lines, and reaches the correct answer. Chain of thought prompting reaches the same answer using three to four times as many tokens. This shows that steering does not make the model imitate a chain of thought style response. Instead, it causes the model to begin deriving an answer rather than stating one directly, and the model stops once it has derived enough to answer.

\begin{algorithm}[t]
\footnotesize
\caption{COMPASS}
\label{alg:compass}
\begin{algorithmic}[1]
\Require frozen $W_O^{l,h}, W_U$; pool $\mathcal{D}$; $K$
\State \textbf{One graded pass} over $\mathcal{D}$: label $y_i$, activations $a_l^h(q_i)$, first generated token $t(q_i)$ \hfill $\forall\, q_i \in \mathcal{D}$
\State Compute $\theta_l^h,\ \sigma_l^h$ by Eqs.~\eqref{eq:theta},~\eqref{eq:sigma}; \quad $\delta_l^h \gets W_O^{l,h}\sigma_l^h\theta_l^h$ \hfill $\forall\, (l,h)$
\State $p_+, p_- \gets$ first generated token histograms over correct, incorrect rows
\State $u \gets W_U^{\top}(p_+ - p_-)$
\State $s_l^h \gets \langle u, \delta_l^h \rangle$; \quad $\mathcal{S} \gets$ top-$K$ heads by $s_l^h$
\State Apply Eq.~\eqref{eq:iti} at test time, with $\theta_l^h \equiv 0\ \forall\, h \notin \mathcal{S}$
\end{algorithmic}
\end{algorithm}



\begin{figure}[t]
\centering
\begin{minipage}[t]{0.49\linewidth}\vspace{0pt}
\begin{tcolorbox}[colback=white,colframe=gray!55,boxrule=0.5pt,arc=2pt,left=4pt,right=4pt,top=3pt,bottom=3pt,fontupper=\scriptsize,equal height group=figexamples,valign=top]
\textbf{Llama-3.1-8B, GSM8K.} \emph{Mary has 30 sheep. She gets 1 kg of milk from half of them and 2 kg of milk from the other half every day. How much milk does she collect every day?} \hfill gold: 45

\medskip
\rowtag{basegold}{DIRECT}\; 15 kg. \hfill\mbox{\badmark\ \ntok{3}}

\medskip
\rowtag{steerblue}{COMPASS}\; To find the total milk collected, we need to calculate the milk from the first half and the second half separately and then add them together. First half: 15 sheep * 1 kg = 15 kg. Second half: 15 sheep * 2 kg = 30 kg. Total milk collected = 15 kg + 30 kg = 45 kg. \hfill\mbox{\okmark\ \ntok{74}}

\medskip
\rowtag{cotgray}{COT}\; \#\# Step 1: Calculate the number of sheep that produce 1 kg of milk per day. Mary has 30 sheep, and half of them produce 1 kg of milk per day. So, the number of sheep that produce 1 kg of milk per day is 30 / 2 = 15. \#\# Step 2: Calculate the number of sheep that produce 2 kg of milk per day. [\ldots\ Steps 3--5 \ldots] Therefore, the final answer is: $\boxed{45}$. \hfill\mbox{\okmark\ \ntok{282}}
\end{tcolorbox}
\end{minipage}\hfill
\begin{minipage}[t]{0.49\linewidth}\vspace{0pt}
\begin{tcolorbox}[colback=white,colframe=gray!55,boxrule=0.5pt,arc=2pt,left=4pt,right=4pt,top=3pt,bottom=3pt,fontupper=\scriptsize,equal height group=figexamples,valign=top]
\textbf{Qwen3-4B, MATH-500.} \emph{In how many ways can 6 girls and 2 boys sit in a row if the 2 boys insist on sitting next to each other?} \hfill gold: 10{,}080

\medskip
\rowtag{basegold}{DIRECT}\; Final answer: 864. \hfill\mbox{\badmark\ \ntok{8}}

\medskip
\rowtag{steerblue}{COMPASS}\; To solve this, treat the 2 boys as a single unit. Then we have 7 units to arrange (6 girls + 1 boy unit). These 7 units can be arranged in $7!$ ways. The 2 boys can be arranged among themselves in $2!$ ways. So the total number of arrangements is: $7!\times 2! = 5040\times 2 = 10080$. Final answer: 10080. \hfill\mbox{\okmark\ \ntok{108}}

\medskip
\rowtag{cotgray}{COT}\; We are given: 6 girls, 2 boys. We are to find the number of ways they can sit in a row if the 2 boys must sit next to each other. \#\#\# Step 1: Treat the 2 boys as a single unit. Since the 2 boys must sit next to each other, we can treat them as one ``block''. This block can be in two different internal arrangements [\ldots\ Steps 2--4 \ldots] Final Answer: $\boxed{10080}$. \hfill\mbox{\okmark\ \ntok{323}}
\end{tcolorbox}
\end{minipage}
\caption{\textbf{Steering elicits a short derivation, not a chain-of-thought transcript.} The direct-answer prompt gets the answer wrong. With COMPASS, the same model reaches the correct answer with a short derivation, generating $3.0$--$3.8\times$ lesser tokens than a chain-of-thought prompt.}
\label{fig:examples}
\end{figure}

\section{Experiments and Results}
\label{sec:experiments}

We evaluate COMPASS across three LLMs over mathematical reasoning benchmarks, measuring accuracy ($\uparrow$) against ground-truth answers, transfer of fitted directions to unseen datasets, and ablations isolating which components of the method drive performance.

\noindent \textbf{Datasets.} We use GSM8K~\cite{cobbe2021gsm8k}, SVAMP~\cite{patel2021svamp}, and MATH-500~\cite{lightman2023verify} for main experiments, with disjoint train and test splits. We evaluate zero-shot transfer, without re-fitting, on GSM-Plus~\cite{li2024gsmplus} and HARP~\cite{yue2024harp}. All benchmarks are short-answer mathematics with a single gold answer per problem, graded by exact or numeric equivalence. Full details are provided in Appendix~\ref{app:splits}.

\noindent \textbf{Baselines.} We compare against (i) Inference-Time Intervention (ITI)~\cite{li2023iti}, which selects heads by probe accuracy and steers the activation them along the probe direction, and (ii) Fractional Reasoning~\cite{liu2025fractional}, which steers every layer of the LLM along a pre-computed reasoning direction from contrastive prompts. Full baseline details are provided in Appendix~\ref{app:baselines}.

\noindent \textbf{Models.} We apply all methods to three open-weight models namely Llama-3.1-8B-Instruct~\cite{grattafiori2024llama3}, Qwen3-4B~\cite{yang2025qwen3}, and Gemma4-12B-it~\cite{gemmateam2026gemma4}.

\noindent \textbf{Oracles.} For reference only, we report a CoT oracle (the same model prompted for explicit chain-of-thought) and a frontier oracle, GPT-5.4-mini with reasoning effort \emph{high}. All methods are evaluated and graded identically.

\noindent \textbf{Basic Setup.} For every sample, we use a fixed, LLM specific brief answer prompt that instructs the model to produce a short answer without explanation. Full prompt details are in Appendix~\ref{app:prompts}. On the training split, we run this prompt once per sample and store the activation of every attention head at the last prompt token, prior to generation. Each generation is then graded against the ground truth answer, labeling the sample correct or incorrect. Full grading details are in Appendix~\ref{app:grading}. For every head, we obtain a correctness direction as the difference between the mean activation over correct samples and the mean activation over incorrect samples. $\alpha$ is a hyperparameter controlling the intensity of the intervention, and we select $\alpha$ separately for every dataset through a hyperparameter search. Evaluation is a single zero shot pass on the test split of each dataset.

\noindent \textbf{Head and Layer Selection.} $s_l^h$ ranks heads reliably within a layer, but not across layers at different depths, since it does not account for how much a head's contribution is transformed before reaching the unembedding layer (Appendix~\ref{app:operating-point}). We therefore restrict head selection to a contiguous band of layers, chosen using the layer aggregated \emph{elicitation mass}, $m_l = \sum_h \max(s_l^h, 0)$, computed by summing only the positive head scores within each layer. Full head and band selection details are in Appendix~\ref{app:operating-point}. Within the selected band, the $K=8$ highest scoring heads are steered.

\subsection{COMPASS Predominantly Outperforms Steering Baselines}
\label{sec:main-results}
Table~\ref{tab:main} reports accuracy and mean generated tokens for the direct answer baseline (standard prompt), ITI~\cite{li2023iti}, Fractional Reasoning~\cite{liu2025fractional}, and COMPASS, with the CoT and frontier model (GPT-5.4 mini) reported as reference. COMPASS improves over the direct answer baseline on every model and dataset, and over both ITI and Fractional Reasoning on every model and dataset except Gemma4-12B on SVAMP, where Fractional Reasoning performs best. On GSM8K, COMPASS closes most of the gap to the CoT reference on Llama and Qwen3, reaching 83.7\% against an 83.4\% ceiling and 89.8\% against 92.3\% respectively, and recovers a substantial part of the gap on Gemma4-12B, reaching 88.0\% against a 96.1\% ceiling. It does so while generating 20 to 70\% fewer tokens than CoT across datasets, for instance 68 versus 224 tokens on Qwen3 SVAMP. COMPASS thus achieves accuracy comparable to or approaching explicit reasoning at a fraction of its token cost.

\begin{table}[t]
\centering
\footnotesize
\setlength{\tabcolsep}{3pt}
\caption{Accuracy (\%) scores with mean generated tokens in gray. The heads selected by COMPASS lift the direct-answer (standard prompt) baseline toward the chain-of-thought and frontier oracles at a fraction of their token cost. ITI and Fractional Reasoning were re-run and we report their optimal performance.}
\label{tab:main}
\resizebox{\linewidth}{!}{%
\begin{tabular}{l ccc ccc ccc}
\toprule
& \multicolumn{3}{c}{\textbf{Llama-3.1-8B}} & \multicolumn{3}{c}{\textbf{Qwen3-4B}} & \multicolumn{3}{c}{\textbf{Gemma4-12B}} \\
\cmidrule(lr){2-4} \cmidrule(lr){5-7} \cmidrule(lr){8-10}
& GSM8K & SVAMP & MATH-500 & GSM8K & SVAMP & MATH-500 & GSM8K & SVAMP & MATH-500 \\
\midrule
Standard Prompt
& 66.8 \tok{93} & 81.0 \tok{48} & 42.4 \tok{345}
& 76.2 \tok{68} & 81.0 \tok{22} & 55.2 \tok{250}
& 49.4 \tok{67} & 74.7 \tok{3} & 68.4 \tok{347}\\
ITI~\cite{li2023iti}
& 73.8 \tok{132} & 68.7 \tok{45} & 23.6 \tok{409}
& 71.9 \tok{45} & 44.3 \tok{78} & 30.6 \tok{56}
& 22.3 \tok{52} & 64 \tok{46} & 30.3 \tok{872}\\
Fractional Reasoning~\cite{liu2025fractional}
& 79.9 \tok{214} & 81.0 \tok{119} & 42.2 \tok{524}
& 88.3 \tok{220} & 89.3 \tok{118} & 77.8 \tok{561}
& 65.4 \tok{55} & \textbf{85.6} \tok{7} & 72.7 \tok{371}\\
\rowcolor{steer}
\textbf{COMPASS (ours)}
& \textbf{83.7} \tok{143} & \textbf{85.7} \tok{81} & \textbf{45.2} \tok{515}
& \textbf{89.8} \tok{138} & \textbf{90.7} \tok{68} & \textbf{77.8} \tok{404}
& \textbf{88.0} \tok{140} & 79.3 \tok{51}& \textbf{79.8} \tok{366}\\
\midrule
CoT \textcolor{gray}{(oracle)}
& 83.4 \tok{205} & 85.0 \tok{157} & 47.0 \tok{788}
& 92.3 \tok{303} & 91.7 \tok{224} & 83.2 \tok{924}
& 96.1 \tok{338} & 94.3 \tok{261} & 91.4 \tok{730}\\
GPT-5.4-mini \textcolor{gray}{(oracle)}
& 95.1 \tok{273} & 94.7 \tok{202} & 88.2 \tok{786}
& 95.1 \tok{273} & 94.7 \tok{202} & 88.2 \tok{786}
& 95.1 \tok{273} & 94.7 \tok{202} & 88.2 \tok{786} \\
\bottomrule
\end{tabular}%
}
\end{table}

\subsection{COMPASS Transfers Without Refitting to Unseen Datasets}
\label{sec:transfer}
Table~\ref{tab:transfer} reports accuracy on GSM-Plus and HARP using the direction, heads, and $\alpha$ obtained from the training splits of GSM8K and MATH-500 respectively, applied without modification. On GSM-Plus, accuracy rises from 60.4\% to 71.1\% on Llama and from 65.6\% to 80.4\% on Qwen3. On HARP, accuracy increases from 21.5\% to 22.7\% on Llama and from 35.4\% to 49.1\% on Qwen3. GSM-Plus consists of adversarial modifications of GSM8K test problems, and HARP is a harder competition level dataset than MATH-500. Since the directions and heads were never exposed to either dataset, this transfer indicates that they capture a property of the model rather than a fit to the source dataset.

\begin{table}[!t]
\centering
\footnotesize
\setlength{\tabcolsep}{3pt}
\caption{Transfer of steering heads and strength selected on GSM8K (to GSM-Plus) and MATH (to HARP), applied zero-shot, without re-tuning. Accuracy (\%) with mean generated tokens in gray.}
\label{tab:transfer}
\resizebox{0.6\linewidth}{!}{%
\begin{tabular}{l cc cc}
\toprule
& \multicolumn{2}{c}{\textbf{Llama-3.1-8B}} & \multicolumn{2}{c}{\textbf{Qwen3-4B}} \\
\cmidrule(lr){2-3} \cmidrule(lr){4-5}
& GSM-Plus & HARP & GSM-Plus & HARP \\
\midrule
Standard Prompt
& 60.4 \tok{136} & 21.5 \tok{366}
& 65.6 \tok{82} & 35.4 \tok{396} \\
\rowcolor{steer}
\textbf{COMPASS (ours)}
& \textbf{71.1} \tok{180} & \textbf{22.7} \tok{750}
& \textbf{80.4} \tok{164} & \textbf{49.1} \tok{575} \\
\midrule
CoT \textcolor{gray}{(oracle)}
& 72.7 \tok{247} & 27.1 \tok{1020}
& 84.1 \tok{360} & 56.9 \tok{1505} \\
GPT-5.4-mini \textcolor{gray}{(oracle)}
& 88.4 \tok{385} & 69.7 \tok{1509}
& 88.4 \tok{385} & 69.7 \tok{1509} \\
\bottomrule
\end{tabular}%
}
\end{table}

\subsection{Ablations Studies}
\label{sec:ablations}
We ablate three components of COMPASS in isolation, the layer band, the head set, and the steering direction, to identify which are necessary for the observed performance improvements.

\begin{figure}[h]
\centering
\includegraphics[width=\linewidth]{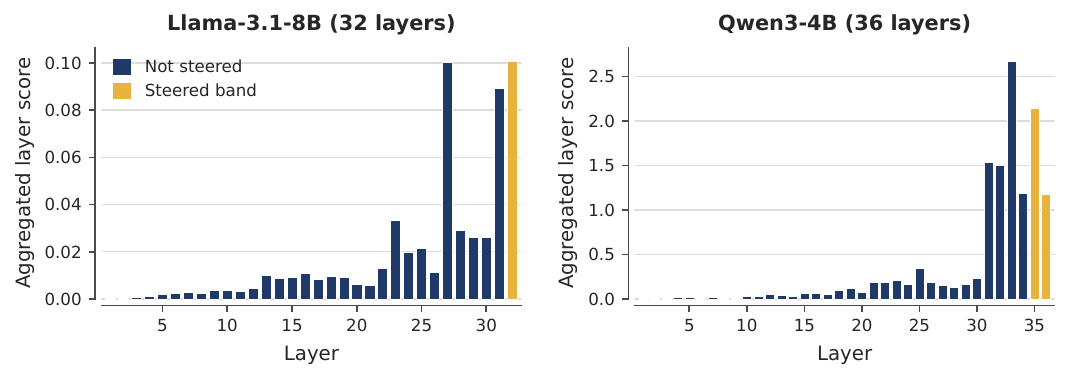}
\caption{\textbf{Aggregated layer score} The sum of the positive head-attribution scores in each layer. Yellow marks the layers from which the steering heads are selected based on the score (Eq~\ref{eq:score}).}
\label{fig:layer-mass}
\end{figure}

\noindent \textbf{Layer Band Selection is Concentrated in the Final Layers.} Each head is scored by the shift it delivers toward the target outcome according to (Eq.~\ref{eq:score}), and a layer's aggregated score is $m_l$. Figure~\ref{fig:layer-mass} plots $m_l$ for both LLama-3.1-8B-it and Qwen3-4B. For roughly the first two thirds of the network, $m_l$ is near zero, indicating that heads in these layers contribute little to the output. $m_l$ then rises sharply over the last few layers, motivating restriction of head selection to this late band rather than the full network.

\begin{figure}[h]
\centering
\includegraphics[width=\linewidth]{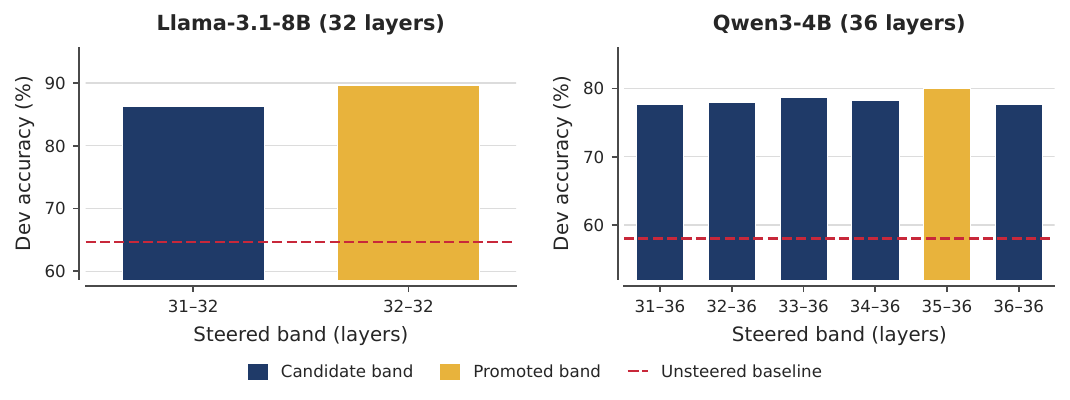}
\caption{\textbf{Best development set accuracy per candidate band.} Bands are compared at constant $\alpha$ and the comparison isolates \emph{which} heads are steered rather than how hard. Yellow is the band the scan promoted (Llama 32--32 at $\alpha=15.66$; Qwen3 35--36 at $\alpha=25.53$). The dashed line is the unsteered direct-answer baseline on the same rows.}
\label{fig:band-accuracy}
\end{figure}

We validate this choice on a held out, development split containing 300 samples, disjoint from both the training and the test set (Appendix~\ref{app:splits}). Holding $\alpha$ fixed, we steer using each candidate band so that only the head set varies. Figure~\ref{fig:band-accuracy} shows that the band with the highest $m_l$ achieves the best or near best development accuracy on both models, confirming that $m_l$ identifies an effective intervention band.

\noindent \textbf{Both Head Selection and the Correctness Direction are Necessary.} Within the selected band, we compare three settings on the full MATH-500 test set, holding the band and $\alpha$ fixed:
\begin{enumerate}
  \item \textbf{Selected heads, correctness direction}: our method, top $K{=}8$ heads by $s_l^h$, steered along the correctness direction $\theta_l^h$.
  \item \textbf{Random heads, correctness direction}: $K{=}8$ heads drawn uniformly at random from the same band, steered along $\theta_l^h$.
  \item \textbf{Selected heads, random direction}: the same top $K{=}8$ heads, steered along a random unit vector of matched magnitude.
\end{enumerate}

\begin{figure}[t]
\centering
\begin{minipage}[t]{0.325\linewidth}\centering
\includegraphics[width=\linewidth]{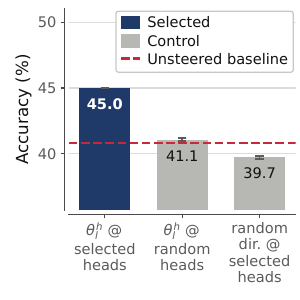}
\end{minipage}\hfill
\begin{minipage}[t]{0.325\linewidth}\centering
\includegraphics[width=\linewidth]{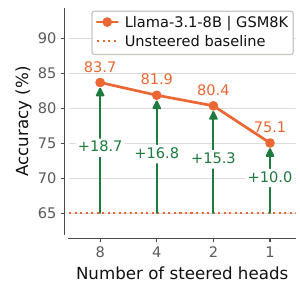}
\end{minipage}\hfill
\begin{minipage}[t]{0.325\linewidth}\centering
\includegraphics[width=\linewidth]{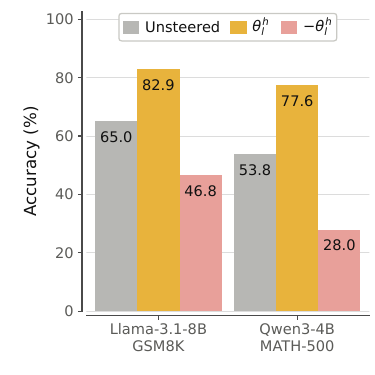}
\end{minipage}
\caption{\textbf{Head-level ablations.} Left: Head and direction controls, on the full 500-row split. $K$ and $\alpha$ are identical across bars. Only the intervention
differs (error bars: sample standard deviation).
Middle: Accuracy as only the $K$ best-ranked heads are steered, with band and $\alpha$ frozen.
Right: Reverses the sign of the injected direction at the
selected heads.}
\label{fig:ablation-heads}
\end{figure}

Figure~\ref{fig:ablation-heads} (left) shows that our method reaches 45.0\% accuracy, a $+4.2$ point gain over the 40.8\% baseline. Randomizing head selection while keeping the correctness direction reduces this to 41.1\%, within one third of a point of baseline. Randomizing the direction while keeping the selected heads is actively harmful, at 39.7\%, below baseline. Both the head set and the correctness direction are therefore required for the effect and neither alone is sufficient.

\noindent \textbf{The Effect Concentrates in Remarkably Few Heads.} Figure~\ref{fig:ablation-heads} (middle) varies $K \in \{8,4,2,1\}$ on Llama-3.1-8B GSM8K. Accuracy degrades gradually as $K$ decreases: 83.7\%, 81.9\%, 80.4\%, and 75.1\% at $K{=}1$, each still a substantial gain over the 65.0\% baseline. A single head retains $+10.0$ of the $+18.7$ point gain achieved at $K{=}8$, indicating that most of the effect is concentrated in a small number of heads rather than distributed evenly across the selected set.

\noindent \textbf{Reversing the Correctness Direction Actively Harms Accuracy.} Figure~\ref{fig:ablation-heads} (right) reverses the sign of the steering direction at the selected heads, injecting $-\theta_l^h$ in place of $\theta_l^h$, with the band, $\alpha$, and head set otherwise unchanged. On Llama-3.1-8B GSM8K, accuracy falls from 82.9\% at $\theta_l^h$ to 46.8\% at $-\theta_l^h$, below the 65.0\% unsteered baseline. On Qwen3-4B MATH-500, accuracy falls from 77.6\% to 28.0\%, below the 53.8\% baseline. Reversal does not merely remove the gain, it drives the model below its unsteered accuracy, confirming that $\theta_l^h$ is signed toward correctness rather than an arbitrary axis of variation the model happens to be sensitive to.

\section{Conclusions and Future Work}
\label{sec:conclusions}

We presented COMPASS, an inference-time steering framework that recovers a reasoning intervention from outcome supervision alone, without chain of thought demonstrations or a predefined notion of reasoning. A single forward pass over a pool of graded problems yields a per-head correctness direction and a logit-space attribution score, together identifying the few heads whose contribution shifts the first generated token toward a correct answer. Across three model families and multiple math benchmarks, COMPASS outperforms existing activation-steering baselines, closes much of the gap to explicit chain of thought at a fraction of the token cost, and transfers to unseen datasets without re-fitting.

While effective in this setting, COMPASS opens several directions for future work. Our attribution score is defined for domains with a binary notion of correctness, such as short-answer math questions. Extending it to open-ended tasks such as proof generation, where outcomes are not directly gradable, is a natural next step. Similarly, while our ablations identify which heads causally drive correctness, a mechanistic account of why these specific heads emerge, and why they promote tokens that initiate correct derivations, is left to future work. We view COMPASS as a step toward using a model's own outcomes, rather than externally imposed notions of reasoning, to identify and steer the internal computation that produces them.

\begin{ack}
This work was performed under the auspices of the U.S.\ Department of Energy (DOE) by Lawrence Livermore National Laboratory under Contract No. DE-AC52-07NA27344. This work was supported by LDRD 25-SI-001. Portions of this work were performed in part at VISLab, UC Riverside. LLNL-CONF-2024134.
\end{ack}

\bibliographystyle{unsrt}
\bibliography{references}


\newpage
\appendix
\section{Appendix for COMPASS: Finding Where Reasoning Lives in Language Models}
This section contains detailed explanations of every protocol followed in our experimental setup. 
Our code is publicly available at : \href{https://github.com/pratyay2510/COMPASS.git}{https://github.com/pratyay2510/COMPASS.git} 
\subsection{Prompts}
\label{app:prompts}

\definecolor{llamabox}{RGB}{232,240,254}
\definecolor{llamaframe}{RGB}{31,58,104}
\definecolor{qwenbox}{RGB}{235,244,235}
\definecolor{qwenframe}{RGB}{31,122,62}
\definecolor{gemmabox}{RGB}{255,244,229}
\definecolor{gemmaframe}{RGB}{176,101,20}

\newtcolorbox{promptbox}[3][]{colback=#2,colframe=#3,boxrule=0.6pt,arc=2pt,
  left=6pt,right=6pt,top=4pt,bottom=4pt,
  fonttitle=\sffamily\bfseries\footnotesize,fontupper=\ttfamily\footnotesize,
  coltitle=white,colbacktitle=#3,toptitle=2pt,bottomtitle=2pt,
  before upper=\setlength{\parskip}{5pt},#1}

Llama-3.1-8B and Qwen3-4B use one prompt on every dataset. Gemma-4-12B-it uses two prompts, both with thinking mode off. On GSM8K and SVAMP it uses a direct answer prompt. On MATH-500 the prompt asks for only the final answer.

The prompt is deliberately terse and asks for the answer without an explanation, so that the baseline reflects what the model does when reasoning is not requested. Each box below reproduces the full prompt string exactly as the code builds it. The marker \texttt{\{problem\}} shows where the question text is inserted, and a blank line inside a box is a blank line in the prompt. The system message is empty for every family, so the rendered input is a single user turn.

\begin{promptbox}[title=Llama-3.1-8B \quad all datasets]{llamabox}{llamaframe}
You are a helpful assitant. Question: \{problem\}. Give me the answer directly.
\end{promptbox}

\begin{promptbox}[title=Qwen3-4B \quad all datasets]{qwenbox}{qwenframe}
Answer the following question with only the final answer. Do not show any working or explanation.

Question: \{problem\}

Final answer:
\end{promptbox}

\begin{promptbox}[title=Gemma-4-12B-it \quad GSM8K and SVAMP]{gemmabox}{gemmaframe}
Answer the following question directly.

Question: \{problem\}
\end{promptbox}

\begin{promptbox}[title=Gemma-4-12B-it \quad MATH and MATH-500]{gemmabox}{gemmaframe}
Answer the following question with only the final answer.

Question: \{problem\}
\end{promptbox}

Decoding is greedy everywhere, with a cap of 512 new tokens on the GSM8K and SVAMP fitting pools and 2{,}048 new tokens on MATH and on every evaluation split. A generation that hits the cap is counted incorrect.

\subsection{Grading}
\label{app:grading}

\definecolor{fmtbox}{RGB}{242,242,240}
\definecolor{fmtframe}{RGB}{91,91,88}

Free form generations do not reliably carry a marker such as \verb|\boxed{}|. A correct answer that ends in ``$\ldots = \frac{109}{33}$'' is extracted as ``33'' by a naive last number rule. Grading therefore proceeds in two passes over every generation.

The first pass is deterministic. An answer is extracted from the raw generation and compared with the gold answer through a fixed cascade. The cascade checks normalized string equality, then numeric equality at relative tolerance $10^{-6}$ with percent signs and unit tails stripped, then reduction of a derivation chain to its final value, and finally bounded symbolic equivalence through SymPy. We call the resulting label the single pass grade.

The second pass follows Fractional Reasoning~\citep{liu2025fractional} and uses an LLM formatter. A second call to the same model, never steered, restates the generation as a single line of the form \texttt{Answer: \textbackslash[\textbackslash boxed\{\ldots\}\textbackslash]}, and the same deterministic cascade then grades the boxed answer. The system message of the formatting call tells the model that its job is to extract the final short answer from the more detailed answer. The user prompt depends on the model and the dataset as follows.

Every model uses the prompt below on every dataset, The placeholder \texttt{\{reasoning\}} carries the generation being graded.

\begin{promptbox}[title=Formatting prompt]{fmtbox}{fmtframe}
Here is a worked solution:

\{reasoning\}

Copy its final answer, exactly as the solution states it, into this line and output nothing else:

Answer: \textbackslash[ \textbackslash boxed\{your\_answer\_here\} \textbackslash]

Do not solve the problem. Do not explain. Do not recompute anything. If the solution states no final answer, box its last expression.
  \end{promptbox}

The final label of a row is the OR of the single pass grade and the formatted answer grade. Only rows the single pass grade marks incorrect are sent to the formatting call, so the formatter can promote a row whose answer the extraction missed but can never demote a row the deterministic cascade already accepts. This is due to the fact that the formatting call is itself a generation and can fail as one. Hence, a regrade that can only add keeps such failures from destroying correct labels. Truncated generations are sent to the formatter as well, since a capped generation can still contain its final answer, and their label follows the same rule.

The same instrument grades everything, so every accuracy score comes from a single universal grader. Accuracy is always reported over all rows of the split.

\subsection{Dataset Splits}
\label{app:splits}

Table~\ref{tab:datasets} shows the datasets that have been used for this work and their corresponding details. 

\begin{table}[h]
\centering
\footnotesize
\caption{Benchmarks. Train splits are fitting pools only; test splits are evaluated in full. GSM-Plus and HARP are used for transfer only.}
\label{tab:datasets}
\begin{tabular}{p{0.2\linewidth} p{0.74\linewidth}}
\toprule
\textbf{Dataset} & \textbf{Description} \\
\midrule
GSM8K~\cite{cobbe2021gsm8k} & Grade-school word problems with integer answers; 7{,}473 train and 1{,}319 test problems. \\
\midrule
SVAMP~\cite{patel2021svamp} & 1{,}000 one-step arithmetic word problems built by perturbing existing problems so that surface cues do not determine the operation; 700/300 train/test split of the public release (ChilleD/SVAMP). Train and test share no question text. \\
\midrule
MATH-500~\cite{lightman2023verify} & The 500-problem subset of the MATH test set~\cite{hendrycks2021math}, seven subjects (prealgebra to precalculus) at five difficulty levels, LaTeX answers. No fitting is done on these 500 problems: the Llama selection is fitted on the 7{,}500-problem MATH train split, and the Qwen3 selection on the remaining 4{,}500 MATH test problems (MATH test minus MATH-500, matched on normalized problem text and verified disjoint). \\
\midrule
GSM-Plus~\cite{li2024gsmplus} & 10{,}552 adversarial perturbations of GSM8K test problems across eight perturbation types. The 1{,}319 \emph{critical-thinking} rows are unanswerable by construction (gold ``None'') and are dropped, leaving 9{,}233 evaluated problems. \\
\midrule
HARP~\cite{yue2024harp} & 4{,}780 short-answer problems from US national mathematics competitions (AMC, AIME, and others), six difficulty levels, LaTeX answers. No train split by design; steered with the MATH selection. \\
\bottomrule
\end{tabular}
\end{table}

Every dataset uses three disjoint row sets, and the convention is the same for all of them: a \emph{fitting pool}, from which the correctness direction and the head ranking are estimated; a 300-row \emph{dev} (scan) set, on which the depth band and the strength $\alpha$ are chosen; and the \emph{test} split, which is generated once at the promoted configuration and never consulted before that.

The details of the three fully disjoint sets are as follows:

\paragraph{Fitting Set}
This is the provided train set for each dataset. It is collected once under the direct-answer prompt (Section~\ref{sec:experiments}). Each row stores the generation, its grade against gold, and the activation of every attention head at the last prompt token. The pool supplies both the outcome target $u$ and the per-head mass-mean directions $\theta_l^h$ with their scales $\sigma_l^h$. The convention differs slightly between GSM8K, MATH and SVAMP. Details are as follows:
\begin{itemize}
    \item \textbf{GSM8K and MATH}: Both GSM8K and MATH have large training sets (GSM8K - 7473 problems, MATH - 7500 problems). For them, after collecting the entire training sets, we construct this fitting pool of $1600$ rows where the correct-incorrect split is exactly equal (800-800). This is done so that the correctness direction remains unbiased and no class imbalance affects either the $\theta_l^h$ or $s_l^h$.
    \item \textbf{SVAMP}: The training set of SVAMP has only $700$ problems. In this dataset, we use $400$ rows as our fitting set. The rest $300$ are used as the dev set. 
\end{itemize}

\paragraph{Dev (scan) set}
For all datasets, this is a set of $300$ rows. For GSM8K and MATH, due to its large size, it is a random set of rows from the spare ones after the fitting pool is identified. For SVAMP, it is carved out of the training set before any fitting takes place. Throughout the study, the scan set is always fully disjoint from the fitting set. 

\paragraph{Test Set}
This is the benchmark's evaluation split, taken in full. It is fully disjoint from both the fitting and dev sets.

\paragraph{GSM-Plus and HARP (transfer).}
Neither dataset has a fitting pool or a dev set. The direction, the head set, and $\alpha$ selected on GSM8K are applied unchanged to GSM-Plus. Those selected on MATH are applied unchanged to HARP. GSM-Plus released $10,552$ problems. However, $1,319$ problems are \emph{critical-thinking} problems, which are unanswerable by construction (gold ``None''). Hence, GSM-Plus is evaluated on $9,233$ problems. HARP is evaluated on all $4,780$ rows.

\subsection{The Operating Point: Depth Band and Dose}
\label{app:operating-point}

The elicitation mass $m_l = \sum_h \max(s_l^h, 0)$,  gives a per-layer profile of how much correct-opening logit the layer can deliver. Only positive scores are summed: a head whose injection moves the logits away from a correct opening is not a candidate, and letting it cancel a useful head in the same layer would misreport what that layer is worth. The profile is essentially flat through the first two thirds of the network, then a sharp rise over the last four to six layers (Figure~\ref{fig:layer-mass}).

This makes a band, the natural search space. We take the \emph{elbow}, $\ell^*$ to be the lowest layer $l$ such that it is the deepest point from which the mass never again falls below 35\% of its peak.
\[
m_{l'} \ge 0.35\ \times \max (m), \forall     l' \ge l.  
\]
The candidate layers are the nested bands $[\ell^\star,L],[\ell^\star{+}1,L],\ldots,[L,L]$. Within a band, the $K=8$ highest-scoring heads are steered.

\paragraph{Matched dose and the operating point.}
Because $\alpha$ multiplies a per-head norm, the same $\alpha$ is a different perturbation in every band. Bands are therefore compared at matched \emph{dose}, $D=\alpha\sum_{(l,h)\in\mathcal S}\|\delta_l^h\|$, the total norm added to the residual stream by the selected set $\mathcal S$. The widest band fixes a ladder of four to five doses from an $\alpha$ range ($\alpha\in[8,32]$). Every other band receives the $\alpha$ values that land it on the same doses. These sets of points (band, $\alpha$) are used to steer the dev sets of each dataset. The single operating point (band, $\alpha$) with the highest dev accuracy is promoted, and is used to steer the test set. 

\subsection{Baselines and Oracles}
\label{app:baselines}
ITI and Fractional Reasoning are re-implemented and run on our models, prompts, splits, token caps, and grader. 

ITI selects the $K=8$ heads with the highest probe validation accuracy over all heads of the model, steers them along the same mass-mean direction, and is swept over a fixed $\alpha$ ladder ($\alpha\in[8,32]$) and reported at the best accuracy.  

Fractional Reasoning experiments are constructed using its own codebase. It builds a steering vector from the mean and first principal component of the (chain-of-thought $-$ direct) last-token differences over 256 training problems. Intervention is applied at every layer with norm-preserving rescaling. The paper does not specify a gold $\alpha$, but reports a band ($\alpha\in[0.01,0.2]$). We run it on that $\alpha$ range in steps of $0.02$ and report the best achieved accuracy. 

The \emph{CoT oracle} is each model's published chain-of-thought evaluation prompt, run on the full test split at a 2{,}048-token cap and graded identically. This is the reference ceiling for a non-thinking model of that size. It is the accuracy the model reaches when reasoning is explicitly requested. The CoT prompts for the three models are as follows:


\begin{promptbox}[title=Llama-3.1-8B \quad chain of thought]{llamabox}{llamaframe}
Solve the following math problem efficiently and clearly:

- For simple problems (2 steps or fewer):\\{}
Provide a concise solution with minimal explanation.

- For complex problems (3 steps or more):\\{}
Use this step-by-step format:

\#\# Step 1: [Concise description]\\{}
[Brief explanation and calculations]

\#\# Step 2: [Concise description]\\{}
[Brief explanation and calculations]

...

Regardless of the approach, always conclude with:

Therefore, the final answer is: \$\textbackslash boxed\{answer\}\$. I hope it is correct.

Where [answer] is just the final number or expression that solves the problem.

Problem: \{problem\}
\end{promptbox}

\begin{promptbox}[title=Qwen3-4B \quad chain of thought]{qwenbox}{qwenframe}
\{problem\}

Please reason step by step, and put your final answer within \textbackslash boxed\{\}.
\end{promptbox}

\begin{promptbox}[title=Gemma-4-12B-it \quad chain of thought]{gemmabox}{gemmaframe}
\{problem\}

Please reason step by step, and put your final answer within \textbackslash boxed\{\}.
\end{promptbox}

As with the direct answer prompts, \texttt{\{problem\}} marks where the question text is inserted, a blank line inside a box is a blank line in the prompt, and the system message is empty for every family.


\end{document}